\documentclass{article} 
\usepackage{iclr2016_workshop,times}
\usepackage{hyperref}
\usepackage{url}
\usepackage{graphicx}
\usepackage{amsmath}
\usepackage{booktabs}
\usepackage{pbox}
\usepackage{float}

\title{Hardware-oriented Approximation of Convolutional Neural Networks}

\author{Philipp Gysel, Mohammad Motamedi \& Soheil Ghiasi\\
Department of Electrical and Computer Engineering\\
University of California, Davis\\
Davis, CA 95616, USA \\
\texttt{\{pmgysel,mmotamedi,ghiasi\}@ucdavis.edu}}

\begin{document}

\maketitle

\begin{abstract}
High computational complexity hinders the widespread usage of Convolutional Neural Networks (CNNs), especially in mobile devices. Hardware accelerators are arguably the most promising
approach for reducing both execution time and power consumption. One of the most important steps in accelerator development is hardware-oriented model approximation.
In this paper we present Ristretto, a model approximation framework that analyzes
a given CNN with respect to numerical resolution used in representing weights and outputs of convolutional and fully connected layers. Ristretto can condense models by using fixed point arithmetic
and representation instead of floating point.
Moreover, Ristretto fine-tunes the resulting fixed point network. Given a maximum error tolerance of 1\%, Ristretto can successfully condense CaffeNet and 
SqueezeNet to 8-bit. The code for Ristretto is available.

\end{abstract}

\section{Introduction}

The annually held ILSVRC competition has seen state-of-the-art classification accuracies by deep networks such as AlexNet by \cite{krizhevsky2012imagenet}, VGG by \cite{Simonyan15},
GoogleNet \citep{szegedy2015going} and ResNet \citep{he2015deep}. These networks contain
millions of parameters and require billions of arithmetic operations.

Various solutions have been offered to reduce the resource-requirement of CNNs. Fixed point arithmetic is less resource hungry compared to floating point. Moreover, it has been shown
that fixed point arithmetic is adequate for neural network computation \citep{hammerstrom1990vlsi}.
This observation has been leveraged recently to condense deep CNNs. \cite{gupta2015deep} show that networks on datasets like CIFAR-10 (10 images classes)
can be trained in 16-bit.
Further trimming of the same network uses as low as 7-bit multipliers \citep{courbariaux2014low}. Another approach by \cite{DBLP:journals/corr/CourbariauxB16} uses binary weights and activations,
again on the same network.

The complexity of deep CNNs can be split into two parts. First, the convolutional layers contain more than 90\% of the required arithmetic operations. By turning these floating point operations
into operations with small fixed point numbers, both the chip area and energy consumption can be significantly reduced. The second resource-intense layer type are fully connected layers, which
contain over 90\% of the network parameters. As a nice by-product of using bit-width reduced fixed point numbers, the data transfer to off-chip memory
is reduced for fully connected layers. In this paper, we concentrate on approximating convolutional and fully connected layers only.
Using fixed point arithmetic is a hardware-friendly way of approximating CNNs. It allows the use of smaller processing elements and reduces the memory requirements without adding any computational
overhead such as decompression.

Even though it has been shown that CNNs perform well with small fixed point numbers, there exists no thorough investigation of the delicate trade-off between bit-width reduction and accuracy loss.
In this paper we present Ristretto, which 
automatically finds a perfect balance between the bit-width reduction and the given maximum error tolerance. Ristretto performs a fast and fully automated trimming analysis of any given network. This 
post-training tool can be used for application-specific trimming of neural networks.

\section{Mixed Fixed Point Precision}

In the next two sections we discuss quantization of a floating point CNN to fixed point. Moreover, we explain dynamic fixed point, and show how it can be used to further decrease network
size while maintaining the classification accuracy.

\begin{figure}[H]
\includegraphics[width=0.8\linewidth]{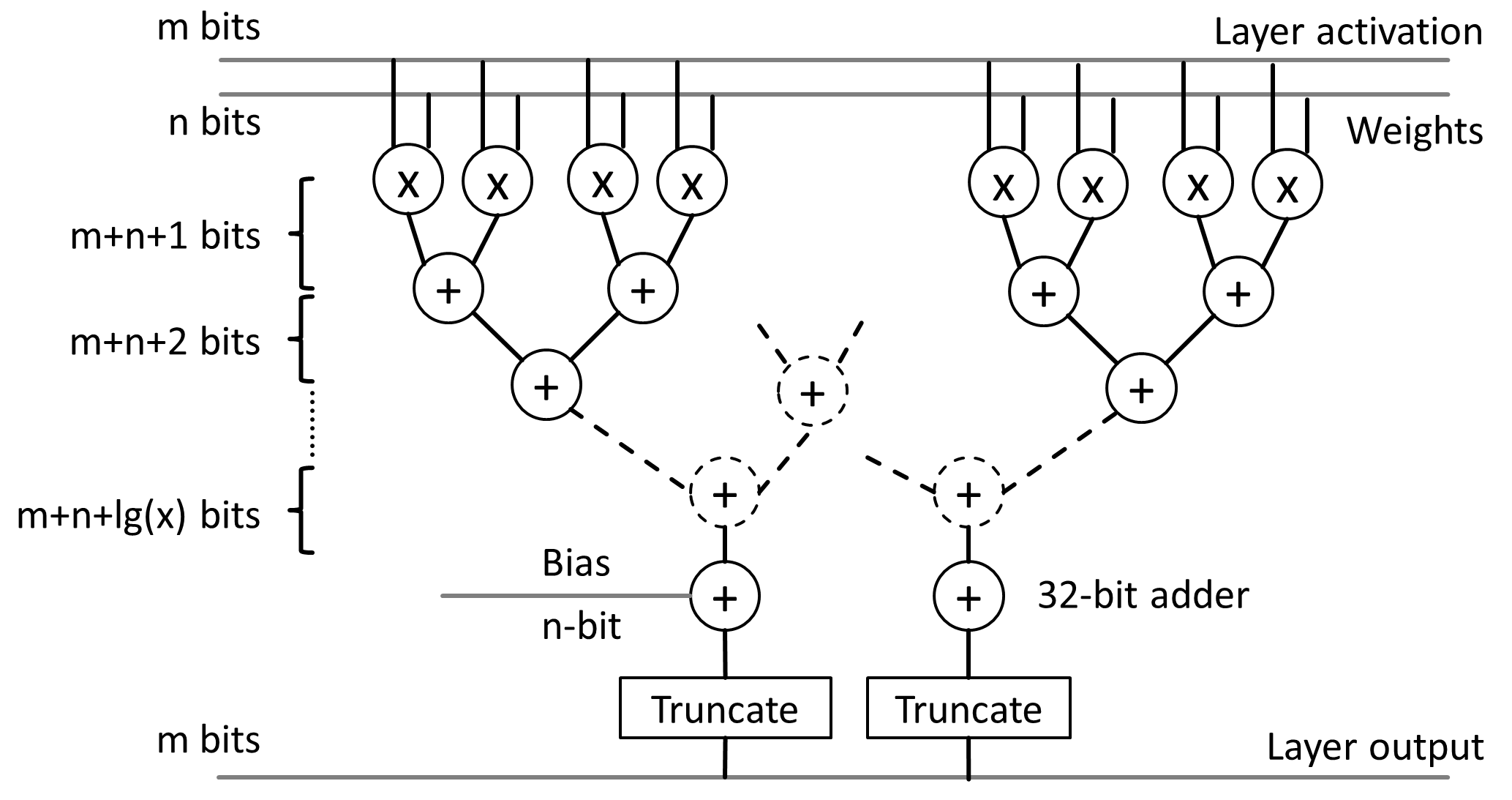}
\centering
\caption{Data path of quantized convolutional and fully connected layers.}
\label{fig:data_flow}
\end{figure}

The data path of fully connected and convolutional layers consists of a series of MAC operations (multiplication and accumulation), as shown in Figure \ref{fig:data_flow}.
The layer activations are multiplied with the network
weights, and the results are accumulated to form the output. As shown by \cite{Qiu:2016:GDE:2847263.2847265}, it is a good approach to use mixed precision, i.e., different
parts of a CNN use different bit-widths.

In Figure \ref{fig:data_flow}, $m$ and $n$ refer to the number of bits for layer outputs and layer weights, respectively.
Multiplication results are accumulated using an adder tree which gets thicker towards the end. The adder outputs in the first level are $m+n+2$ bits wide, and the bit-width grows by 1 bit in each level.
In the last level, the bit-width is $m+n+\lg_{2}x$, where
$x$ is the number of multiplication operations per output value. In the last stage, the bias is added to form the layer output. For each network layer, we need to find the right balance
between reducing the bit-widths ($m$ and $n$) and maintaining a good classification accuracy.

\section{Dynamic Fixed Point}

The different parts of a CNN have a significant dynamic range. In large layers, the outputs are the result of thousands of accumulations, thus the network parameters are much smaller
than the layer outputs. Fixed point has only limited capability to cover a wide dynamic range. Dynamic fixed
point \citep{williamson1991dynamically,courbariaux2014low} is a solution to this problem.

In dynamic fixed point, each number is represented as follows: ${(-1)^s \cdot 2^{-fl} \cdot \sum_{i=0}^{B-2} 2^i \cdot x_i}$. 
Here $B$ denotes the bit-width, $s$ the sign bit, $fl$ is the fractional length, and $x$ the mantissa bits.
The intermediate values in a network have different ranges. Therefor it is desirable to assign fixed point numbers into \textit{groups with constant fl}, such that the number of bits
allocated to the fractional part
is constant within that group. Each network layer is split into two groups: one for the layer outputs, one for the layer weights. This allows to better
cover the dynamic range of both layer outputs and weights, as weights are normally significantly smaller.
On the hardware side, it is possible to realize dynamic fixed point arithmetic using bit shifters.

Different hardware accelerators for deployment of neural networks have been proposed \citep{motamedi2016design, Qiu:2016:GDE:2847263.2847265, han2016eie}. The first important step in accelerator
design is the compression of the network in question. In the next section we present Ristretto, a tool which can condense any neural network in a fast and automated fashion.

\section{Ristretto: Approximation Framework in Caffe}

\textbf{From Caffe to Ristretto} \\
According to Wikipedia, Ristretto is 'a short shot of espresso coffee made with the normal amount of ground coffee but extracted with about half the amount of water'. Similarly,
our compressor removes the unnecessary parts of a CNN, while making sure the essence -- the ability to predict image classes -- is preserved. With its strong community and fast
training for deep CNNs, Caffe \citep{jia2014caffe} is an excellent framework to build on.

Ristretto takes a trained model as input, and automatically brews a condensed network version. Input and output of Ristretto are a network description file (prototxt) and the network parameters.
Optionally, the quantized network can be fine-tuned with Ristretto. The resulting fixed point model in Caffe-format can then be used for a hardware accelerator.

\begin{figure}[H]
\includegraphics[width=1\linewidth]{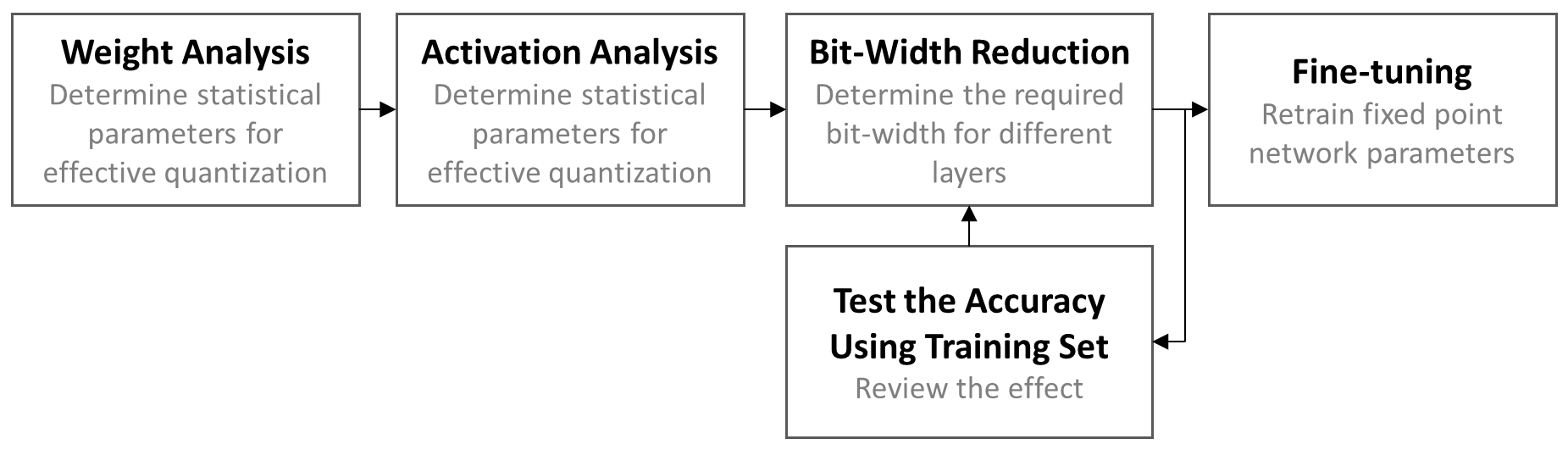}
\centering
\caption{Network approximation flow with Ristretto.}
\label{fig:ristretto}
\end{figure}

\textbf{Quantization flow} \\
Ristretto's quantization flow has five stages (Figure \ref{fig:ristretto}) to compress a floating point network into fixed point.
In the first step, the dynamic range of the weights is analyzed to find a good fixed point representation. For the quantization from floating point to fixed point, we use round-nearest.
The second step runs several thousand images in forward path. The generated layer activations are
analyzed to generate statistical parameters. Ristretto uses enough bits in the integer part of fixed point numbers to avoid saturation of layer activations.
Next Ristretto performs a binary search to find
the optimal number of bits for convolutional weights, fully connected weights, and layer outputs. In this step, a certain network part is quantized, while the rest remains in floating point. Since
there are three network parts that should use independent bit-widths (weights of convolutional and fully connected layers as well as layer outputs), iteratively quantizing one network part
allows us to find the optimal bit-width for each part. Once a good trade-off between small number representation and classification accuracy is found, the resulting fixed point network is retrained.

\textbf{Fine-tuning} \\
In order to make up for the accuracy drop incurred by quantization, the fixed point network is fine-tuned in Ristretto. During this retraining procedure, the network learns how to classify images
with fixed point parameters. Since the network weights can only have discrete values, the main challenge consists in the weight update. We adopt the idea of previous work
\citep{courbariaux2015binaryconnect} which uses \textit{full precision shadow weights}. Small weight updates $\Delta w$ are applied to the full precision weights $w$, whereas the discrete weights $w'$
are sampled from
the full precision weights. The sampling during fine-tuning is done with stochastic rounding. This rounding scheme was successfully used by \cite{gupta2015deep} for weight updates of 16-bit fixed point
networks. 

\begin{figure}[H]
\includegraphics[width=0.8\linewidth]{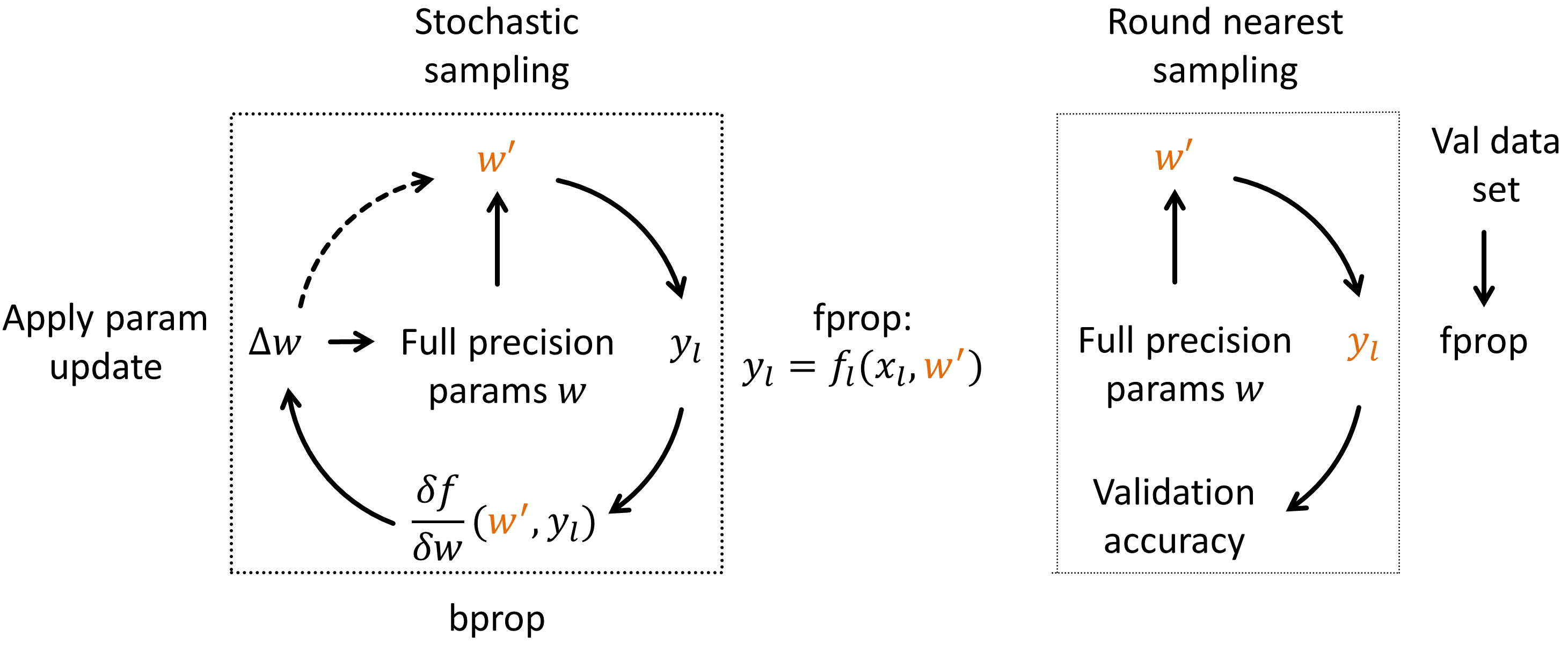}
\centering
\caption{Fine-tuning with shadow weights. The left side shows the training process with full-precision shadow weights. On the right side the fine-tuned network is benchmarked on the validation 
data set. Fixed point values are represented in orange.}
\label{fig:shadow_weights}
\end{figure}

Ristretto uses the fine-tuning procedure illustrated in Figure \ref{fig:shadow_weights}. For each batch, the full precision weights are quantized to fixed point. During forward propagation,
these discrete weights are used to compute the layer outputs $y_l$. Each layer $l$ turns its input batch $x_l$ into output $y_l$, according to its function $f_l: (x_l,w') \rightarrow y_l$. Assuming
the last layer computes the loss, we denote $f$ as the overall CNN function.

The goal of back propagation is to compute the error gradient $\delta f/\delta w$ with respect to each fixed point parameter. For
parameter updates
we use the Adam rule by \cite{kingma2014adam}. As an important observation, we do \textit{not} quantize layer outputs to fixed point during fine-tuning. We use floating point layer outputs
instead, which enables Ristretto to analytically compute the error gradient with respect to each parameter. In contrast, the validation of the network is done with fixed point layer outputs.

To achieve the best fine-tuning results, we used a learning rate that is an order of magnitude lower than the last full precision training iteration.
Since the choice of hyper parameters for retraining is crucial \citep{bergstra2012random}, Ristretto
relies on minimal human intervention in this step.

\textbf{Fast fine-tuning with fixed point parameters} \\
Ristretto brews a condensed network with fixed point weights and fixed point layer activations. For simulation of the forward propagation in hardware, Ristretto uses full floating point for accumulation.
This follows the thought of \cite{gupta2015deep} and is conform with our description of the forward data path in hardware (Figure \ref{fig:ristretto}). During fine-tuning, the full precision weights need
to be converted
to fixed point for each batch, but after that all computation can be done in floating point (Figure \ref{fig:shadow_weights}). Therefore Ristretto can fully leverage optimized matrix-matrix
multiplication routines for both forward and backward propagation. \\
Thanks to its fast implementation on the GPU, a fixed point
CaffeNet can be tested on the ILSVRC 2014 validation dataset (50k images) in less than 2 minutes (using one Tesla K-40 GPU).

\section{Results}
\label{sec:results}

In this section we present the results of approximating 32-bit floating point networks by condensed fixed point models. All classification accuracies were obtained running the respective network on
the whole validation dataset. We present approximation results of Ristretto for five different networks. First, we consider LeNet \citep{lecun1998gradient} which can classify handwritten digits
(MNIST dataset).
Second, CIFAR-10 Full model provided by
Caffe is used to classify images into 10 different classes. Third, we condense CaffeNet, which is the Caffe version of AlexNet and classifies images into the 1000 ImageNet categories. Fourth, we
use the BVLC version of GoogLeNet \citep{szegedy2015going} to classify images of the same data set. Finally, we approximate SqueezeNet \citep{SqueezeNet}, a recently proposed architecture with the classification accuracy
of AlexNet, but $>$50X fewer parameters.

\begin{figure}[H]
\includegraphics[width=0.8\linewidth]{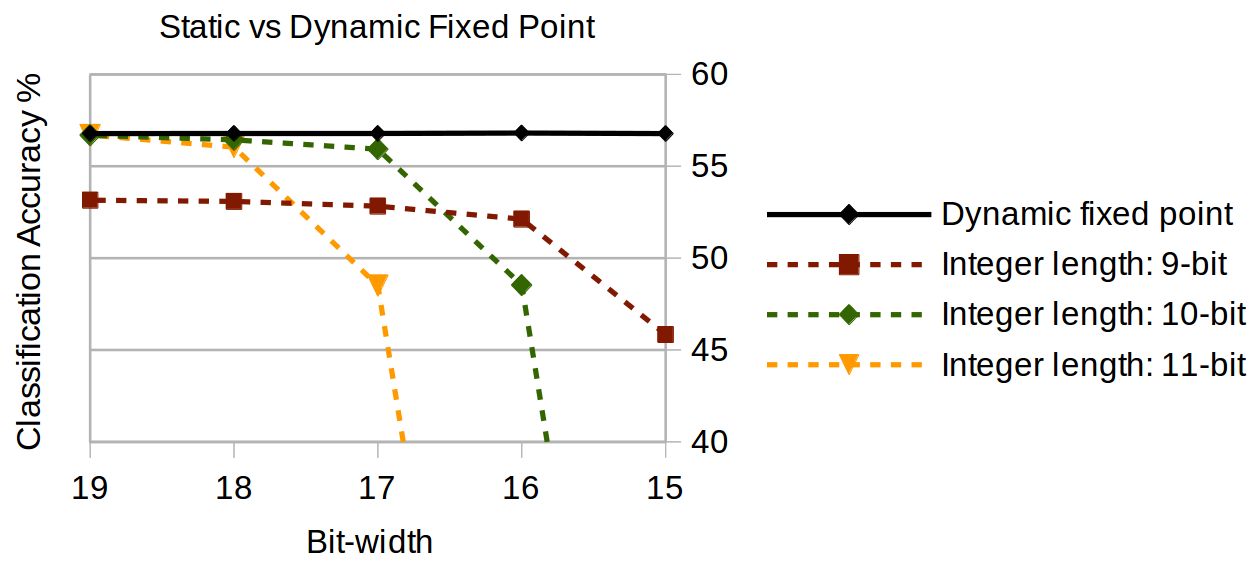}
\centering
\caption{Impact of dynamic fixed point: The figure shows top-1 accuracy for CaffeNet on ILSVRC 2014 validation dataset. Integer length refers to the number of bits assigned to the integer part of
fixed point numbers.}
\label{fig:static_vs_dynamic_fixed_point}
\end{figure}

\textbf{Impact of dynamic fixed point} \\
We used Ristretto to quantize CaffeNet (AlexNet) into fixed point, and
compare traditional fixed point with dynamic fixed point. To allow a simpler comparison, all layer outputs and network parameters share the
same bit-width. Results show a good performance of static
fixed point for as low as 18-bit (Figure \ref{fig:static_vs_dynamic_fixed_point}). However, when reducing the bit-width further, the accuracy starts to drop significantly, while dynamic fixed
point has a stable accuracy.

We can conclude that dynamic fixed point performs significantly better for such a large network. With dynamic fixed point,
we can adapt the number of bits allocated to integer and fractional part, according to the dynamic range of different parts of the network. We will therefore concentrate on dynamic fixed point
for the subsequent experiments.

\textbf{Quantization of individual network parts}\\
In this section, we analyze the impact of quantization on different parts of a floating point CNN.
Table \ref{tab:compress_parts} shows the classification accuracy when the layer outputs,
the convolution kernels or the parameters of fully connected layers are quantized to dynamic fixed point. 

In all three nets, the convolution kernels and layer activations can be trimmed to 8-bit
with an absolute
accuracy change of only 0.3\%. Fully connected layers are more affected from trimming to 8-bit weights, the absolute change is maximally 0.9\%.
Interestingly, LeNet weights can be trimmed to as low as 2-bit, with absolute accuracy change below 0.4\%.

\begin{table}[H]
\caption{Quantization results for different parts of three networks. Only one number category is cast to fixed point, and the remaining numbers are in floating point format.}
\label{tab:compress_parts}
\begin{center}
\begin{tabular}{*5l}
\toprule
Fixed point bit-width & 16-bit & 8-bit & 4-bit & 2-bit \\
\midrule
\multicolumn{5}{l}{\textbf{LeNet, 32-bit floating point accuracy: 99.1\%}} \\
\midrule
Layer output & 99.1\% & 99.1\% & 98.9\% & 85.9\%\\
CONV parameters & 99.1\% & 99.1\% & 99.1\% & 98.9\%\\
FC parameters & 99.1\% & 99.1\% & 98.9\% & 98.7\%\\
\midrule
\multicolumn{5}{l}{\textbf{Full CIFAR-10, 32-bit floating point accuracy: 81.7\%}} \\
\midrule
Layer output & 81.6\% & 81.6\% & 79.6\% & 48.0\%\\
CONV parameters & 81.7\% & 81.4\% & 75.9\% & 19.1\%\\
FC parameters & 81.7\% & 80.8\% & 79.9\% & 77.5\%\\
\midrule
\multicolumn{5}{l}{\textbf{CaffeNet top-1, 32-bit floating point accuracy: 56.9\%}} \\
\midrule
Layer output & 56.8\% & 56.7\% & 06.0\% & 00.1\% \\
CONV parameters & 56.9\% & 56.7\% & 00.1\% & 00.1\% \\
FC parameters & 56.9\% & 56.3\% & 00.1\% & 00.1\% \\
\bottomrule
\end{tabular}
\end{center}
\end{table}

\textbf{Fine-tuning of all considered network parts} \\
Here we report the accuracy of five networks that were condensed and fine-tuned with Ristretto. All networks use dynamic fixed point parameters as well as dynamic fixed point layer outputs for
convolutional and fully connected layers.
LeNet performs well in 2/4-bit, while CIFAR-10 and the three ImageNet CNNs can be trimmed to 8-bit (see Table \ref{tab:compress_all}).
Surprisingly, these compressed networks still perform nearly as well as their floating point baseline. The relative accuracy drops of LeNet, CIFAR-10 and SqueezeNet are very small ($<$0.6\%),
whereas the approximation of the larger CaffeNet and GoogLeNet incurs a slightly higher cost (0.9\% and 2.3\% respectively). We hope we will further improve the fine-tuning results of these larger
networks in the future.

The SqueezeNet architecture was developed by \cite{SqueezeNet} with the goal of a small CNN that performs well on the ImageNet data set. Ristretto can make the already small network even smaller, so that its 
parameter size is less than 2 MB. This condensed network is well-suited for deployment in smart mobile systems.

All five 32-bit floating point networks can be approximated well in 8-bit and 4-bit fixed point. For a hardware implementation,
this reduces the size of multiplication units by about one
order of magnitude. Moreover, the required memory bandwidth is reduced by 4--8X. Finally, it helps to hold 4--8X more parameters in on-chip buffers. The code for reproducing the quantization and
fine-tuning results is available\footnote[1]{\url{https://github.com/pmgysel/caffe}}.

\begin{table}[H]
\caption{Fine-tuned networks with dynamic fixed point parameters and outputs for convolutional and fully connected layers. The numbers in brackets indicate accuracy without fine-tuning.}
\label{tab:compress_all}
\begin{center}
\begin{tabular}{*6l}
\toprule
{} & \pbox{20cm}{Layer \\ outputs} & \pbox{20cm}{CONV \\ parameters} & \pbox{20cm}{FC \\ parameters} & \pbox{20cm}{32-bit floating \\ point baseline} & \pbox{20cm}{Fixed point \\ accuracy} \\
\midrule
LeNet (Exp 1) & 4-bit & 4-bit & 4-bit & 99.1\% & 99.0\% (98.7\%) \\
LeNet (Exp 2) & 4-bit & 2-bit & 2-bit & 99.1\% & 98.8\% (98.0\%) \\
Full CIFAR-10 & 8-bit & 8-bit & 8-bit & 81.7\% & 81.4\% (80.6\%) \\
SqueezeNet top-1 & 8-bit & 8-bit & 8-bit & 57.7\% & 57.1\% (55.2\%) \\
CaffeNet top-1 & 8-bit & 8-bit & 8-bit & 56.9\% & 56.0\% (55.8\%) \\
GoogLeNet top-1 & 8-bit & 8-bit & 8-bit & 68.9\% & 66.6\% (66.1\%) \\
\bottomrule
\end{tabular}
\end{center}
\end{table}

A previous work by \cite{courbariaux2014low} concentrates on training with limited numerical precision.
They can train a dynamic fixed point network on the MNIST data set using just 7-bits to represent activations and weights.
Ristretto doesn't reduce the resource requirements for training, but concentrates on inference instead. Ristretto can produce a LeNet network with 
2-bit parameters and 4-bit activations. Our approach is different in that we train with high numerical precision, then quantize
to fixed point, and finally fine-tune the fixed point network.

Other works \citep{DBLP:journals/corr/CourbariauxB16, rastegari2016xnor} can reduce the bit-width even further to as low as 1-bit, using more advanced
number encodings than dynamic fixed point.
Ristretto's strength lies in its capability to approximate a large number of existing floating point models on challenging data sets. For the five considered networks, Ristretto can quantize activations and weights to 8-bit or lower, at an accuracy drop below 2.3\%, compared to the floating point baseline.

While more sophisticated data compression schemes could be used to achieve higher network size reduction, our approach is very hardware friendly and imposes no additional
overhead such as decompression.

\section{Conclusion and Future Work}

In this work we presented Ristretto, a Caffe-based approximation framework for deep convolutional neural networks. The framework reduces
the memory requirements, area for processing elements and overall power consumption for hardware accelerators.
A large net like CaffeNet can be quantized to 8-bit for both weights and layer outputs while keeping the network's
accuracy change below 1\% compared to its 32-bit floating point counterpart. Ristretto is both fast and automated, and we release the code as an open source project.

Ristretto is in its first development stage. We consider adding new features in the future: 1. Shared weights: Fetching cookbook indices from off-chip
memory, instead of real values \citep{han2015deep}. 2. Network pruning as shown by the same authors. 3. Network binarization as shown by \cite{DBLP:journals/corr/CourbariauxB16} and
\cite{rastegari2016xnor}.
These additional features will help to reduce the bit-width even further, and to reduce the computational complexity of trimmed networks.

\bibliography{iclr2016_workshop}

\begin{thebibliography}{20}
\providecommand{\natexlab}[1]{#1}
\providecommand{\url}[1]{\texttt{#1}}
\expandafter\ifx\csname urlstyle\endcsname\relax
  \providecommand{\doi}[1]{doi: #1}\else
  \providecommand{\doi}{doi: \begingroup \urlstyle{rm}\Url}\fi

\bibitem[Bergstra \& Bengio(2012)Bergstra and Bengio]{bergstra2012random}
Bergstra, J. and Bengio, Y.
\newblock Random {S}earch for {H}yper-{P}arameter {O}ptimization.
\newblock \emph{The Journal of Machine Learning Research}, 13\penalty0
  (1):\penalty0 281--305, 2012.

\bibitem[Courbariaux et~al.(2014)Courbariaux, David, and
  Bengio]{courbariaux2014low}
Courbariaux, M., David, J.-P., and Bengio, Y.
\newblock Training {D}eep {N}eural {N}etworks with {L}ow {P}recision
  {M}ultiplications.
\newblock \emph{arXiv preprint arXiv:1412.7024}, 2014.

\bibitem[Courbariaux et~al.(2015)Courbariaux, Bengio, and
  David]{courbariaux2015binaryconnect}
Courbariaux, M., Bengio, Y., and David, J.-P.
\newblock Binary{C}onnect: {T}raining {D}eep {N}eural {N}etworks with binary
  weights during propagations.
\newblock In \emph{Advances in {N}eural {I}nformation {P}rocessing {S}ystems},
  pp.\  3105--3113, 2015.

\bibitem[Courbariaux et~al.(2016)Courbariaux, Hubara, Soudry, El-Yaniv, and
  Bengio]{DBLP:journals/corr/CourbariauxB16}
Courbariaux, M., Hubara, I., Soudry, D., El-Yaniv, R., and Bengio, Y.
\newblock Binarized {N}eural {N}etworks: {T}raining {D}eep {N}eural {N}etworks
  with {W}eights and {A}ctivations {C}onstrained to +1 or -1.
\newblock \emph{arXiv preprint arXiv:1602.02830}, 2016.

\bibitem[Gupta et~al.(2015)Gupta, Agrawal, Gopalakrishnan, and
  Narayanan]{gupta2015deep}
Gupta, S., Agrawal, A., Gopalakrishnan, K., and Narayanan, P.
\newblock Deep {L}earning with {L}imited {N}umerical {P}recision.
\newblock In \emph{Proceedings of the 32nd International Conference on Machine
  Learning (ICML-15)}, pp.\  1737--1746, 2015.

\bibitem[Hammerstrom(1990)]{hammerstrom1990vlsi}
Hammerstrom, D.
\newblock A {VLSI} {A}rchitecture for {H}igh-{P}erformance, {L}ow-{C}ost,
  {O}n-chip {L}earning.
\newblock In \emph{IJCNN International Joint Conference on Neural Networks,
  1990}, pp.\  537--544. IEEE, 1990.

\bibitem[Han et~al.(2016{\natexlab{a}})Han, Liu, Mao, Pu, Pedram, Horowitz, and
  Dally]{han2016eie}
Han, S., Liu, X., Mao, H., Pu, J., Pedram, A., Horowitz, M.~A., and Dally,
  W.~J.
\newblock {EIE}: {E}fficient {I}nference {E}ngine on {C}ompressed {D}eep
  {N}eural {N}etwork.
\newblock \emph{arXiv preprint arXiv:1602.01528}, 2016{\natexlab{a}}.

\bibitem[Han et~al.(2016{\natexlab{b}})Han, Mao, and Dally]{han2015deep}
Han, S., Mao, H., and Dally, W.~J.
\newblock {D}eep {C}ompression: {C}ompressing {D}eep {N}eural {N}etworks with
  {P}runing, {T}rained {Q}uantization and {H}uffman {C}oding.
\newblock In \emph{International Conference on Learning Representations},
  2016{\natexlab{b}}.

\bibitem[He et~al.(2015)He, Zhang, Ren, and Sun]{he2015deep}
He, K., Zhang, X., Ren, S., and Sun, J.
\newblock Deep {R}esidual {L}earning for {I}mage {R}ecognition.
\newblock \emph{arXiv preprint arXiv:1512.03385}, 2015.

\bibitem[Iandola et~al.(2016)Iandola, Moskewicz, Ashraf, Han, Dally, and
  Keutzer]{SqueezeNet}
Iandola, F.~N., Moskewicz, M.~W., Ashraf, K., Han, S., Dally, W.~J., and
  Keutzer, K.
\newblock Squeeze{N}et: {A}lex{N}et-level accuracy with 50x fewer parameters
  and \textless0.5{MB} model size.
\newblock \emph{arXiv:1602.07360}, 2016.

\bibitem[Jia et~al.(2014)Jia, Shelhamer, Donahue, Karayev, Long, Girshick,
  Guadarrama, and Darrell]{jia2014caffe}
Jia, Y., Shelhamer, E., Donahue, J., Karayev, S., Long, J., Girshick, R.,
  Guadarrama, S., and Darrell, T.
\newblock Caffe: {C}onvolutional {A}rchitecture for {F}ast {F}eature
  {E}mbedding.
\newblock In \emph{Proceedings of the ACM International Conference on
  Multimedia}, pp.\  675--678. ACM, 2014.

\bibitem[Kingma \& Ba(2015)Kingma and Ba]{kingma2014adam}
Kingma, D. and Ba, J.
\newblock Adam: {A} {M}ethod for {S}tochastic {O}ptimization.
\newblock In \emph{International Conference on Learning Representations}, 2015.

\bibitem[Krizhevsky et~al.(2012)Krizhevsky, Sutskever, and
  Hinton]{krizhevsky2012imagenet}
Krizhevsky, A., Sutskever, I., and Hinton, G.~E.
\newblock Image{N}et {C}lassification with {D}eep {C}onvolutional {N}eural
  {N}etworks.
\newblock In \emph{Advances in {N}eural {I}nformation {P}rocessing {S}ystems},
  pp.\  1097--1105, 2012.

\bibitem[LeCun et~al.(1998)LeCun, Bottou, Bengio, and
  Haffner]{lecun1998gradient}
LeCun, Y., Bottou, L., Bengio, Y., and Haffner, P.
\newblock Gradient-{B}ased {L}earning {A}pplied to {D}ocument {R}ecognition.
\newblock \emph{Proceedings of the IEEE}, 86\penalty0 (11):\penalty0
  2278--2324, 1998.

\bibitem[Motamedi et~al.(2016)Motamedi, Gysel, Akella, and
  Ghiasi]{motamedi2016design}
Motamedi, M., Gysel, P., Akella, V., and Ghiasi, S.
\newblock Design {S}pace {E}xploration of {FPGA}-{B}ased {D}eep {C}onvolutional
  {N}eural {N}etworks.
\newblock In \emph{2016 21st Asia and South Pacific Design Automation
  Conference (ASP-DAC)}, pp.\  575--580. IEEE, 2016.

\bibitem[Qiu et~al.(2016)Qiu, Wang, Yao, Guo, Li, Zhou, Yu, Tang, Xu, Song,
  Wang, and Yang]{Qiu:2016:GDE:2847263.2847265}
Qiu, J., Wang, J., Yao, S., Guo, K., Li, B., Zhou, E., Yu, J., Tang, T., Xu,
  N., Song, S., Wang, Y., and Yang, H.
\newblock Going {D}eeper with {E}mbedded {FPGA} {P}latform for {C}onvolutional
  {N}eural {N}etwork.
\newblock In \emph{Proceedings of the 2016 ACM/SIGDA International Symposium on
  Field-Programmable Gate Arrays}, pp.\  26--35, 2016.

\bibitem[Rastegari et~al.(2016)Rastegari, Ordonez, Redmon, and
  Farhadi]{rastegari2016xnor}
Rastegari, M., Ordonez, V., Redmon, J., and Farhadi, A.
\newblock {XNOR}-{N}et: {I}mage{N}et {C}lassification {U}sing {B}inary
  {C}onvolutional {N}eural {N}etworks.
\newblock \emph{arXiv preprint arXiv:1603.05279}, 2016.

\bibitem[Simonyan \& Zisserman(2015)Simonyan and Zisserman]{Simonyan15}
Simonyan, K. and Zisserman, A.
\newblock Very {D}eep {C}onvolutional {N}etworks for {L}arge-{S}cale {I}mage
  {R}ecognition.
\newblock In \emph{International Conference on Learning Representations}, 2015.

\bibitem[Szegedy et~al.(2015)Szegedy, Liu, Jia, Sermanet, Reed, Anguelov,
  Erhan, Vanhoucke, and Rabinovich]{szegedy2015going}
Szegedy, C., Liu, W., Jia, Y., Sermanet, P., Reed, S., Anguelov, D., Erhan, D.,
  Vanhoucke, V., and Rabinovich, A.
\newblock Going {D}eeper with {C}onvolutions.
\newblock In \emph{Proceedings of the IEEE Conference on Computer Vision and
  Pattern Recognition}, pp.\  1--9, 2015.

\bibitem[Williamson(1991)]{williamson1991dynamically}
Williamson, D.
\newblock Dynamically scaled fixed point arithmetic.
\newblock In \emph{IEEE Pacific Rim Conference on Communications, Computers and
  Signal Processing, 1991}, pp.\  315--318. IEEE, 1991.

\end{thebibliography}
\bibliographystyle{iclr2016_workshop}

\end{document}